\documentclass[10pt,twocolumn,letterpaper]{article}
\usepackage[pagenumbers]{cvpr} % To force page numbers, e.g. for an arXiv version

\usepackage{graphicx}
\usepackage{amsmath}
\usepackage{amssymb}
\usepackage{booktabs}

\usepackage{amsfonts,amssymb} 
\usepackage{algorithm}
\usepackage{algorithmic}
\usepackage{multirow}
\usepackage{microtype}

\usepackage[pagebackref,breaklinks,colorlinks]{hyperref}

\usepackage[capitalize]{cleveref}
\crefname{section}{Sec.}{Secs.}
\Crefname{section}{Section}{Sections}
\Crefname{table}{Table}{Tables}
\crefname{table}{Tab.}{Tabs.}

\def\cvprPaperID{11785} % *** Enter the CVPR Paper ID here
\def\confName{CVPR}
\def\confYear{2023}

\begin{document}

\title{MetaViewer: Towards A Unified Multi-View Representation}

\author{Ren Wang\\
Shandong University\\
{\tt\small xxlifelover@gmail.com}
\and
Haoliang Sun\\
Shandong University\\
{\tt\small haolsun.cn@gmail.com}\\
\and
Yuling Ma\\
Shandong Jianzhu University\\
{\tt\small mayuling20@sdjzu.edu.cn}
\and
Xiaoming Xi\\
Shandong Jianzhu University\\
{\tt\small fyzq10@126.com}
\and
Yilong Yin\\
Shandong University\\
{\tt\small ylyin@sdu.edu.cn}
}

\maketitle

%%%%%%%%% ABSTRACT
\begin{abstract}
Existing multi-view representation learning methods typically follow a specific-to-uniform pipeline, extracting latent features from each view and then fusing or aligning them to obtain the unified object representation. However, the manually pre-specify fusion functions and view-private redundant information mixed in features potentially degrade the quality of the derived representation. To overcome them, we propose a novel bi-level-optimization-based multi-view learning framework, where the representation is learned in a uniform-to-specific manner. Specifically, we train a meta-learner, namely MetaViewer, to learn fusion and model the view-shared meta representation in outer-level optimization. Start with this meta representation, view-specific base-learners are then required to rapidly reconstruct the corresponding view in inner-level. MetaViewer eventually updates by observing reconstruction processes from uniform to specific over all views, and learns an optimal fusion scheme that separates and filters out view-private information. Extensive experimental results in downstream tasks such as classification and clustering demonstrate the effectiveness of our method. 
\end{abstract}

\section{Introduction}
\label{sec:intro}

Multi-view representation learning mines a unified representation from multiple views of the same entity \cite{DBLP:journals/tkde/LiYZ19, DBLP:journals/pami/ZhangFHCXTX20, DBLP:journals/pami/SunDL21, DBLP:journals/tcsv/ZhengZL22}. Each view acquired by different sensors or sources contains both view-shared consistency information and view-specific information. Among them, view-specific information consists of complementary and redundant components, where the former can be considered as a supplement to the consistency information, while the latter is highly specific and may be adverse for the unified representation \cite{DBLP:conf/iclr/Federici0FKA20, DBLP:conf/aaai/GengHZH21}. Therefore, a high-quality representation is required to retain the consistency and complementary information, as well as filter out the view-private redundant ones \cite{DBLP:conf/cvpr/XuT0P0022}.

\begin{figure}[t]
  \centering
  \includegraphics[width=1.\linewidth]{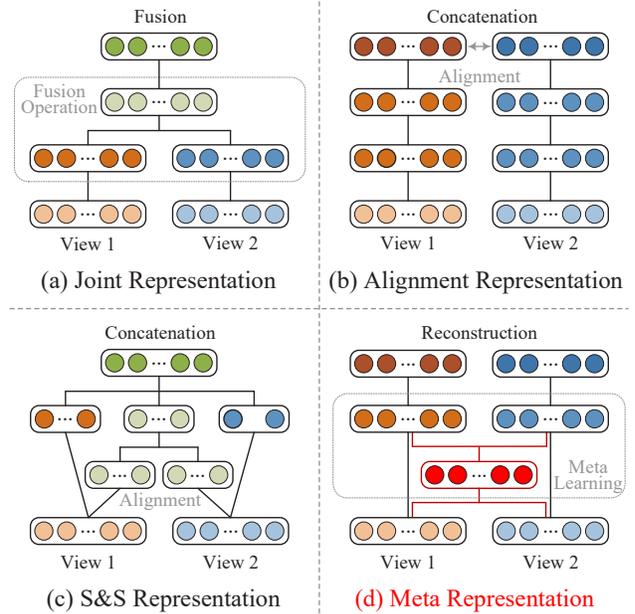} 
  \caption{(a), (b) and (c) show three multi-view learning frameworks following the \textit{specific-to-uniform} pipeline, where the unified representation is obtained by fusing or concatenating view-specific features. (d) illustrates our \textit{uniform-to-specific} manner, where a meta-learner learns to fusion by observing reconstruction from unified representation to specific views.}
  \label{fig:fig1}
  
\end{figure}

Given the data containing two views $x_1$ and $x_2$, the prevailing multi-view representation methods typically follow a \textit{specific-to-uniform} pipeline and can be roughly characterized as:
\begin{equation}
    H := f(x_1;W_f) \circ g(x_2;W_g),
\end{equation}
where $f$ and $g$ are encoding (or embedding \cite{lin2022dual}) functions that map the original view data into the corresponding latent features with the trainable parameters $W_f$ and $W_g$. These latent features are subsequently aggregated into the unified representation using the designed aggregation operator $\circ$. with different aggregation strategies, existing approaches can be further subdivided into the joint, alignment, and a combined share-specific (S$\&$S) representation \cite{DBLP:journals/pami/BaltrusaitisAM19, DBLP:journals/tkde/LiYZ19, DBLP:journals/pami/JiaJZCDCHY21}.

Joint representation focuses on the integration of complementary information by directly fusing latent features, where the $\circ$ is represented as fusion strategies, such as graph-based module \cite{DBLP:journals/tip/TaoHNZY17, DBLP:journals/tip/NieCLL18}, neural network \cite{DBLP:conf/cvpr/McLaughlinRM16, DBLP:conf/cvpr/FeichtenhoferPZ16}, or other elaborate functions \cite{DBLP:journals/jmlr/SrivastavaS14, DBLP:conf/aistats/GunasekarYYC15, DBLP:conf/ijcai/ChengZCLHR16}. While alignment representation seeks to perform alignment between view-specific features to retain the consistency information and specifies $\circ$ as the alignment operator, measured by distance \cite{DBLP:conf/mm/FengWL14, DBLP:conf/mm/LiDLS03}, similarity \cite{DBLP:conf/nips/FromeCSBDRM13, DBLP:journals/pami/KarpathyF17}, or correlation\cite{DBLP:conf/icml/AndrewABL13, DBLP:conf/icml/WangALB15, DBLP:conf/aaai/JingWDSC17}. As a trade-off way, S$\&$S representation explicitly distinguishes latent features into shared and specific representation and only aligns the shared part \cite{DBLP:conf/aaai/XueNWCSY17, DBLP:journals/pami/HuLT18, DBLP:conf/aaai/LuoZZC18, DBLP:journals/pami/JiaJZCDCHY21}. Fig. \ref{fig:fig1} (a) - (c) show the above three branches of the \textit{specific-to-uniform} paradigm.

Despite demonstrating promising results, the aggregation way inherently suffers from potential risks in the following two aspects: (1) The derived unified representation is usually the concatenation or fusion of learned latent features with the manually pre-specify rules. It makes them fairly hard to be generally applied in practice due to the significant variation of fusion schemes relying on the downstream tasks and training views \cite{DBLP:conf/nips/ShuXY0ZXM19}. (2) Even finding a well-performing fusion scheme, view-private redundant information mixed in latent features also degrades the quality of the fused unified representation. Several studies have noticed the second issue and attempted to distinguish the redundant information from view features via multi-level feature modeling \cite{DBLP:conf/cvpr/XuT0P0022} or matrix factorization \cite{DBLP:journals/tnn/ZhaoYZCYW21}. However, recent works indicate the view-specific information could be not automatically separated at feature level \cite{DBLP:journals/jmlr/SalzmannEUD10, DBLP:journals/pami/JiaJZCDCHY21}. In addition, the first issue has received little attention.

In this paper, we propose a novel multi-view representation learning framework based on bi-level optimization meta-learning. In contrast to the \textit{specific-to-uniform} pipeline, our model emphasizes learning unified representation in a \textit{uniform-to-specific} manner, as illustrated in Fig. \ref{fig:fig1} (d). In detail, we build a meta-learner, namely MetaViewer, to learn fusion and model a unified meta representation in outer-level optimization. Based on this meta representation, view-specific base-learners are required to rapidly reconstruct the corresponding view in the inner-level. MetaViewer eventually updates by observing reconstruction processes over all views, thus learning optimal fusion rules to address the first issue. On the other hand, the rapid reconstruction from uniform representation to specific views in inner-level optimization essentially models the information that cannot be fused, i.e., view-private parts, solving the second issue. After alternate training, the resulting meta representation is closer to each view, which indicates that it contains as much view-shared information as possible as well as avoids the hindrance of redundant information. Extensive experiments on multiple benchmarks validate the performance of our MetaViewer. The unified meta representation learned from multiple views achieves comparable performance to state-of-the-art methods in downstream tasks such as clustering and classification. The core contributions of this paper are as follows.

\begin{enumerate}
\item We propose a novel insight for multi-view representation learning, where the unified representation is learned in a \textit{uniform-to-specific} manner.
\item Our MetaViewer achieves the data-driven fusion of view features in meta-learning paradigm. To the best of our knowledge, this could be the first meta-learning-based work in multi-view representation scenarios.
\item MetaViewer decouples the modeling of view-shared and view-private information via bi-level optimization, alleviating the hindrance of redundant information.
\item Extensive experimental results validate the performance of our approach in several downstream tasks.
\end{enumerate}

\section{Related Work}
\label{subsec:2_related}

\subsection{Multi-view learning} \label{subsec:2_1}
Multi-view representation learning is not a new topic and has been widely used in downstream tasks such as retrieval, classification, and clustering. This work focuses on multi-view representation in unsupervised deep learning scope, and related works can be summarized into two main categories \cite{DBLP:journals/ijon/YanHMYY21}. One is the deep extension of traditional methods, where representative ones include deep canonical correlation analysis (DCCA) \cite{DBLP:conf/icml/AndrewABL13} and its variants \cite{DBLP:conf/icml/WangALB15, DBLP:conf/aaai/SunSSL20, DBLP:conf/cvpr/YuanLLQ00G22}. DCCA intends to discover the nonlinear mapping for two views to a common space in which their canonical correlations are maximally preserved. These methods benefit from a sound theoretical foundation, but also usually have strict restrictions on the number and form of views. 

Another alternative is the multi-view deep network. Early deep-based approaches attempted to build different architectures for handling multi-view data, such as CNN-based \cite{DBLP:conf/cvpr/FengZZJG18, DBLP:conf/cvpr/YuMY18, DBLP:journals/tip/SunZLYS21} and GAN-based model \cite{DBLP:conf/cvpr/XueWCW20, DBLP:conf/icml/HassaniA20}, etc. Some recent approaches focus on better parameter constraints using mutual information \cite{DBLP:conf/nips/BachmanHB19, DBLP:conf/iclr/Federici0FKA20}, comparative information \cite{lin2022dual, DBLP:conf/cvpr/YuLH22}, etc. Most of these existing methods follow a \textit{specific-to-uniform} pipeline. In contrast, the underlying assumption of our MetaViewer learns from uniform to specific. The most related work is the MFLVC \cite{DBLP:conf/cvpr/XuT0P0022}, which separates view-private information from latent features at the parameter level. The essential difference is that we model the view-private information in inner-level optimization, allowing outer-level observes the modeling process and future meta learners the optimal fusion scheme.

\begin{figure*}[t]
  \centering
   \includegraphics[width=1.\linewidth]{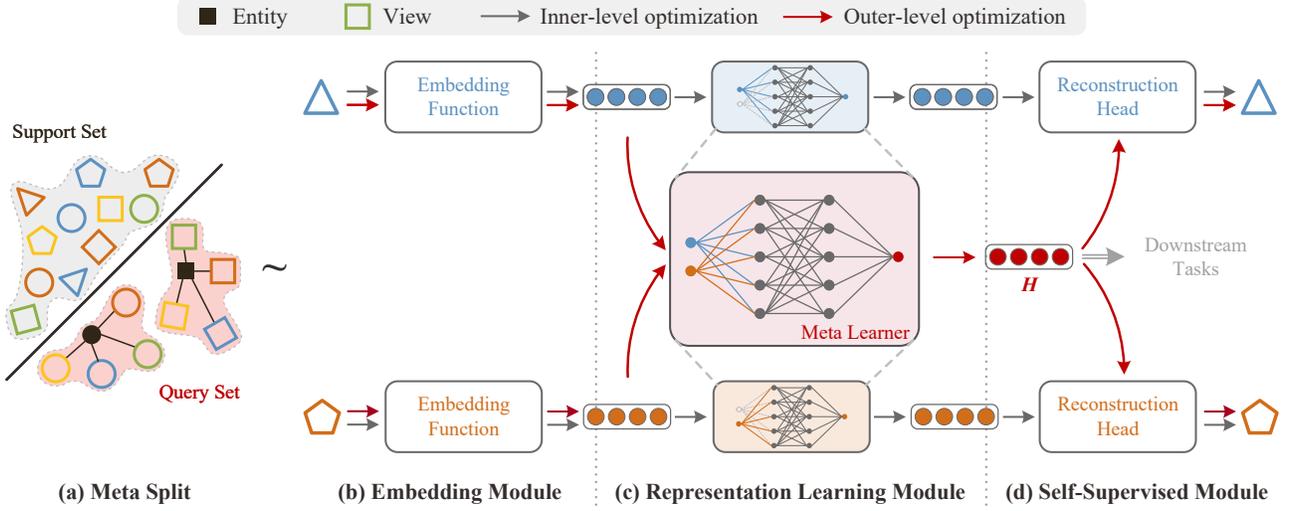}
   \caption{The overall framework of MetaViewer, contains (a) the meta split of multi-view data and three modules (b) embedding module, (c) representation module, and (d) self-supervised module. These modules are trained with a bi-level optimization. Inner-level (dark gray arrows) learns the view-specific reconstruction on the support set, and outer-level (red arrows) updates the entire model to learn the fusion scheme and the unified representation $H$ for downstream tasks by validating on the query set.}
   \label{fig:fig2}
\end{figure*}

\subsection{Meta-learning} \label{subsec:2_2}
Optimization-based meta-learning is a classic application of bi-level optimization designed to learn task-level knowledge to quickly handle new tasks \cite{DBLP:journals/pami/HospedalesAMS22, DBLP:journals/air/HuismanRP21}. A typical work, MAML \cite{DBLP:conf/icml/FinnAL17}, learns a set of initialization parameters to solve different tasks with few steps of updates. Similar meta paradigm has been used to learn other manually designed parts, such as the network structure \cite{DBLP:conf/iccv/LiuMZG0C019}, optimizer\cite{DBLP:conf/icml/WichrowskaMHCDF17, DBLP:conf/iclr/ZhengCHW22}, and even sample weight \cite{DBLP:conf/nips/ShuXY0ZXM19, DBLP:conf/acl/MaoW00X22}. Similarly, we try to meta-learn fusion of multi-view feature for a unified representation. There also exist some works that consider both meta-learning paradigms and multi-view data\cite{DBLP:conf/nips/VuorioSHL19, DBLP:journals/tip/GaoLYY22, DBLP:journals/kbs/MaZWLK22}. However, they are dedicated to exploit the rich information contained in multiple views to improve the performance of the meta-learner on few-shot tasks or self-supervised scenario. Instead, we train a meta-learner to derive the high-quality shared representations from multi-view data via the bi-level optimization. To the best of our knowledge this is the first work to learn multi-view representation with meta-learning paradigm.

\section{MetaViewer}
\label{sec:3_method}
Given a set of unlabeled multi-view dataset $\mathcal{D} = \{x_i\in \mathbb{R}^{d_x}\}_{i=1}^N$, where $N$ are the number of samples and each sample entity $x_i=\{x_i^1, x_i^2, \dots, x_i^v\}_{v=1}^V$ contains $V$ views. The view-incomplete scenario means that partial views of some samples are missing or unavailable, i.e., $\mathcal{D}_{inc} = \{\mathcal{D}_{c}, \{\mathcal{D}_{u}^v\}_{v=1}^V\}$, where $\mathcal{D}_{c}$ and $\mathcal{D}_{u}^v$ indicate subsets of samples with complete views and unavailable $v$-$th$ view, respectively. Our goal is learn a unified high-quality representation $H$ for each entity by observing available views and filtering self-private information as much as possible. The overall framework of our MetaViewer is shown in Fig. \ref{fig:fig2}, including three mainly modules and a bi-level optimization process. The outer-level learns a meta-learner to learn a optimal fusion function and derive the unified representation $H$, and inner-level reconstructs original views from $H$ in few update steps, which explicitly models and separates view-private information and ensure the representation quality. In following subsections, we first introduce the entire structure of the MetaViewer and then elaborate the bi-level optimization process.

\subsection{The entire structure}
\label{subsec:3_1}

\textbf{Embedding module} aims to transform heterogeneous view into the latent feature space, where transformed view embeddings have the same dimension as each other. To this end, we conduct a view-specific embedding function $f_v$ for each view, where $v=1, 2, \dots, V$. Given the $v$-$th$ view data $x^v$ of the entity $x$, the corresponding embedding $z^v$ can be computed by 
\begin{equation}
z^v = f_v(x^v,\phi_{f_v}),
\label{eq:embedding_function}
\end{equation}
where $z^v \in \mathbb{R}^d$ and $f_v$ typically instantiated as a multi-layer neural network with learnable parameters $\phi_{f_v}$. 

\textbf{Representation learning module} maps the obtained embedding to the view representation, consisting of view-specific base-learners ${\{b_v\}_{v=1}^V}$ and a view-shared meta-learner $m$ (i.e., MetaViewer). The former learns representation for each view embedding, while the latter takes all embeddings as input and outputs the unified representation $H$ that is ultimately used for downstream tasks. Meanwhile, base-learners are generally required to be initialized from parameters of the meta-learner to learning view-shared meta representation (see \ref{subsec:3_2}), thus two types learners should be structurally shared rather than individually designed. To meet the above two requirements simultaneously, MetaViewer is implemented as a channel-oriented $1$-$d$ convolutional layer (C-Conv) with a non-linear function (e.g., ReLU\cite{DBLP:journals/jmlr/GlorotBB11}), as shown in Fig. \ref{fig:fig2} (c).

On the one hand, as a meta-learner, we first concatenate embeddings at the channel level, i.e., the number of the channel in the concatenated feature is equal to the number of views, and then train the MetaViewer to learn the fusion of cross-view information $H \in \mathbb{R}^{d_h}$ by
\begin{equation}
H = m(z^{cat}, \omega),
\label{eq:base_learner}
\end{equation}
where $z^{cat} \in \mathbb{R}^{d \times V}$ indicates the concatenated embedding and $\omega$ is the parameter of the MetaViewer. On the other hand, base-learners could be initialized and trained for learning $v$-$th$ representation $h_{base}^v$ via
\begin{equation}
h_{base}^v = b_v(z^v, \theta_{b_v}(\omega_{sub})),
\label{eq:base_learner}
\end{equation}
where $h_{base}^v \in \mathbb{R}^{d_h}$ and $\theta_{b_v}(\omega_{sub})$, or $\theta_{b_v}(\omega)$ for short, means the base-learner's parameter $\theta_{b_v}$ is initialized from the channel(view)-related part $\omega_{sub}$ of the MetaViewer's parameters. Note that this sub-network mechanism also provides a convenient way to handle incomplete views.

\textbf{Self-supervised module} conducts pre-text tasks to provide effectively supervised objects for model training, represented by different heads. Typically, the reconstruction head $r$ achieves the reconstruction object by re-mapping the representation back to the original view space, i.e., 
\begin{equation}
x_{rec}^v = r_v(z_{rec}^v,\phi_{r_v}),
\label{eq:recontruction_head}
\end{equation}
where $x_{rec}^v \in \mathbb{R}^{d_x}$ is the reconstruction result and $r_v$ is the reconstruction function with learnable parameters $\phi_{r_v}$. Expect that, we can also conduct contrastive or correlation head for the meta-learner to mine the associations across views. Similar self-supervised objectives \cite{DBLP:conf/aaai/SunSSL20, DBLP:conf/cvpr/YuanLLQ00G22, lin2022dual, DBLP:conf/cvpr/YuLH22} have been extensively studied in multi-view learning, and this is not the focus of this work.

\subsection{Training via bi-level optimization} 
\label{subsec:3_2}
Now, we have conducted the entire structure, which can be end-to-end trained to derive the unified multi-view representation, even for incomplete views, in the \textit{specific-to-uniform} manner like most existing approaches. However the data-driven fusion and view-private redundant information still cannot be handling well. So we turn to a opposite \textit{uniform-to-specific} way using a bi-level optimization process, inspired by the meta-learning paradigm. Inner-level focus on the training of view-specific modules for corresponding views, and outer-level updates meta-learner to find the optimal fusion rule through observing the learning over all views. Before the detailed description, we introduce a split way of multi-view data in meta-learning style.

\textbf{Meta-split of multi-view data}. 
Consider a batch multi-view samples $\{\mathcal{D}_{batch}^v\}_{v=1}^V$ from $\mathcal{D}$. For bi-level updating, we randomly and proportionally divide it into two disjoint subsets, marked as support set $S$ and query set $Q$, respectively. As shown in \ref{fig:fig2} (a), support set is used in inner-level for leaning view-specific information, thus the sample attributes in it could be ignored. In contrast, query set retains both view and sample attributes for meta-learner training in outer-level optimization. This meta-split that decoupling view from samples can be naturally transferred to the data with incomplete views $\mathcal{D}_{inc} = \{\mathcal{D}_{c}, \{\mathcal{D}_{u}^v\}_{v=1}^V\}$, where subset with incomplete views $\{\mathcal{D}_{u}^v\}_{v=1}^V$ is used as support set, and the complete part $\mathcal{D}_{c}$ is left as query set.

\begin{figure}[t]
  \centering
  \includegraphics[width=1.\linewidth]{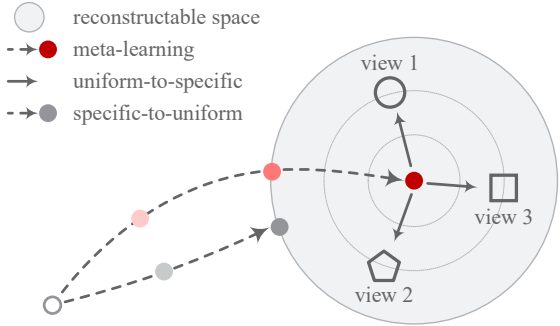}
  \caption{Intuitively, the \textit{specific-to-uniform} observes view features and learns a representation that falls in a reconstructable space. MetaViewer observes the reconstruction process and seeks a unified meta representation that is close as possible to each view.}
  \label{fig:discuss}
\end{figure}

\begin{algorithm}[tb]
\caption{The framework of our MetaViewer.}
\label{alg:algorithm}
\textbf{Require}: Training dataset $\mathcal{D}$, meta parameters $\omega$, base parameters $\{\theta_v\}_{v=1}^V$, view-specific parameters $\{\phi_v\}_{v=1}^V$, the number of view $V$, the iteration step in inner-level optimization $T$.
\begin{algorithmic}[1] %[1] enables line numbers
\STATE Initialize $\omega$, $\{\phi_v\}_{v=1}^V$;
\WHILE{not done} 
    \STATE \# $Outer$-$level$\\
    \STATE Sample and meta-split a batch set from $\mathcal{D}$: \\
    \STATE $\{\mathcal{D}_{batch}^v\}_{v=1}^V = \{S,Q\}$.
    \FOR {$t = 1, \dots, T$}
        \FOR {$v = 1, \dots, V$}
        \STATE \#  $Inner$-$level$\\
        \STATE Initialize $\theta_v = \theta_v(\omega)$, $\tilde{\phi_v} = \phi_v$;
        \STATE Optimize $\theta_v(\omega)$ and $\tilde{\phi_v}$ via Eq. \ref{eq:inner_update}.
        \ENDFOR
    \ENDFOR
    \STATE Optimize $\omega$ via Eq. \ref{eq:outer_update}.
    \STATE Optimize $\{\phi_v\}_{v=1}^V$ via Eq. \ref{eq:outer_update}.
\ENDWHILE
\end{algorithmic}
\end{algorithm}

\textbf{Inner-level optimization}. 
Without loss of generality, take the inner-level update after the $o$-$th$ outer-level optimization as an example. Let $\omega^o$ be the lasted parameters of the MetaViewer, and $\phi_{v}^o = \{\phi_{f_v}^o, \phi_{r_v}^o\}$ denotes the lasted parameters in embedding and self-supervised modules for brevity. We first initial base-learner from meta-learner, i.e., $\theta_{b_v}^{0} = \omega^{o}$, and make a copy of $\phi_{v}^o$ for $\tilde{\phi}_{v}$. Note that the \textit{copy} means gradients with respect to the $\phi_{v}^o$ will not be back-propagated to $\tilde{\phi}_{v}$ and vice versa. Thus, $\theta_{b_v}^{0}$ and $\tilde{\phi}_{v}$ form the initial state for the inner-level optimization. Suppose the $\mathcal{L}_{inner}^v$ is the loss function of the inner-level with respect to the $v$-$th$ view, and then the corresponding update goals are 
\begin{equation}
\theta_{b_v}^{\ast}(\omega), \tilde{\phi}_{v}^{\ast}, = \mathop{\arg\min} \mathcal{L}_{inner}^{v}\left(\theta_{b_v}(\omega^o), \tilde{\phi}_{v}; S^v\right).
\label{eq:inner_update}
\end{equation}
Consider a gradient descent strategy (e.g., SGD \cite{DBLP:conf/compstat/Bottou10}), we can further write the update process of $\theta_{b_v}$:
\begin{equation}
\theta_{b_v}^{i} = \theta_{b_v}^{i-1} - \beta \frac{ \partial \mathcal{L}_{inner}^{v} }{ \partial \theta_{b_v}^{i-1} }, \dots, \theta_{b_v}^{0} = \omega^{o},
\label{eq:base_learner_update}
\end{equation}
where $\beta$ and $i$ denote the learning rate and iterative step of inner-level optimization, respectively. 

\textbf{Outer-level optimization}. 
After several inner-level updates, we obtain a set of optimal view-specific parameters on the support set. Outer-level then updates the meta-learner, embedding, and head modules by training on the query set. With the loss function $\mathcal{L}_{outer}$, the outer-level optimization goal is
\begin{equation}
\omega^{\ast}, \{\phi_{v}^{\ast}\}_1^V = \mathop{\arg\min} \mathcal{L}_{outer}\left(\theta^{*}(\omega), \{\phi_{v}\}_1^V; Q \right).
\label{eq:outer_update}
\end{equation}
By alternately optimizing Eq. \ref{eq:inner_update} and Eq. \ref{eq:outer_update}, we end up with the optimal meta-parameters $\omega^*$ and a set of view-specific parameters $\phi_v^{\ast}$. For a test sample $x_{test}$, its corresponding representation is derived by sequentially feeding the embedding function and the meta-learner. The overall framework of our MetaViewer is shown in Alg. \ref{alg:algorithm}.

\begin{table*}
  \centering
  \begin{tabular}{c|c|c|c|c}%{@{}lc@{}}
    \toprule
    Datasets & \#Views & \#Classes & \#Samples (train, val, test)& View Dimensions \\
    \midrule
    BDGP   & 2 & 5 & 2500 (1500, 500, 500) & 1750; 79 \\
    Handwirtten     & 2 & 10 & 2000 (1200, 400, 400)& 240; 216 \\
    RGB-D   & 2 & 50 & 500 (300, 100, 100) & 12288; 4096 \\
    Fashion-MV     & 3 & 10 & 10000 (6000, 2000, 2000) & 784; 784; 784\\
    MSRA            & 6 & 7 & 210 (126, 42, 42)    & 1302; 48; 512; 100; 256; 210 \\
    Caltech101-20   & 6 & 20 & 2386 (1425, 469, 492)& 48; 40; 254; 1984; 512; 928\\
    \bottomrule
  \end{tabular}
  \caption{The attributes for all datasets used in our experiments.}
  \label{tab:data}
\end{table*}

\begin{table*}
  \centering
  \begin{tabular}{r|c|ccc ccc cc}%{@{}lc@{}}
    \toprule
    Datasets & Metrics  & \makebox[0.08\textwidth][c]{DCCA \cite{DBLP:conf/icml/AndrewABL13}}
                        & \makebox[0.08\textwidth][c]{DCCAE \cite{DBLP:conf/icml/WangALB15}}
                        & \makebox[0.08\textwidth][c]{MIB \cite{DBLP:conf/iclr/Federici0FKA20}}
                        & \makebox[0.08\textwidth][c]{MFLVC \cite{DBLP:conf/cvpr/XuT0P0022}}
                        & \makebox[0.08\textwidth][c]{DCP \cite{lin2022dual}}
                        & \makebox[0.08\textwidth][c]{MVer-R}
                        & \makebox[0.08\textwidth][c]{MVer-C}\\
\midrule
\multirow{3} * {BDGP}          & ACC & 0.4640 & 0.5180 & 0.6940 & \underline{0.8800} & 0.7820 & 0.8280 & \textbf{0.9040}\\
                               & NMI & 0.4163 & 0.5793 & 0.5565 & \underline{0.8397} & 0.7800 & 0.7979 & \textbf{0.8627}\\
                               & ARI & 0.3347 & 0.3208 & 0.4865 & \underline{0.8504} & 0.6725 & 0.6156 & \textbf{0.8925}\\
\midrule 
\multirow{3} * {Handwritten}   & ACC & 0.5725 & 0.6300 & 0.6325 & 0.6400 & 0.6625 & \underline{0.7500} & \textbf{0.8625}\\
                               & NMI & 0.6980 & 0.7504 & 0.6758 & 0.6453 & 0.7056 & \underline{0.7853} & \textbf{0.7896}\\
                               & ARI & 0.5215 & 0.5929 & 0.5216 & 0.4885 & 0.5610 & \underline{0.6721} & \textbf{0.7225}\\
\midrule 
\multirow{3} * {RGB-D}          & ACC & 0.5100 & 0.4800 & 0.5000 & 0.5300 & 0.5200 & \underline{0.5300} & \textbf{0.5700}\\
                               & NMI & 0.8299 & 0.8158 & 0.8113 & \underline{0.8331} & 0.8204 & 0.8241 & \textbf{0.8497}\\
                               & ARI & 0.5202 & 0.4834 & 0.5127 & \underline{0.5407} & 0.5264 & 0.5304 & \textbf{0.5707}\\
\midrule 
\multirow{3} * {Fashion-MV}     & ACC & 0.7070 & 0.7105 & 0.5720 & \underline{0.8320} & 0.6260 & 0.8080 & \textbf{0.8540}\\
                               & NMI & 0.8042 & 0.8112 & 0.7383 & \underline{0.8875} & 0.6838 & 0.8813 & \textbf{0.8876}\\
                               & ARI & 0.6180 & 0.6234 & 0.4762 & \underline{0.7893} & 0.5430 & 0.7505 & \textbf{0.8007}\\
\midrule 
\multirow{3} * {MSRA}          & ACC & 0.3333 & 0.3571 & 0.3095 & \underline{0.7143} & 0.6429 & \textbf{0.7143} & 0.7018\\
                               & NMI & 0.2997 & 0.3285 & 0.2471 & 0.6796 & \underline{0.6801} & \textbf{0.7007} & 0.6126\\
                               & ARI & 0.3341 & 0.3471 & 0.3035 & \underline{0.6891} & 0.6308 & \textbf{0.6893} & 0.6029\\
\midrule 
\multirow{3} * {Caltech101-20} & ACC & 0.3862 & 0.3659 & 0.3598 & 0.3659 & 0.3679 & \underline{0.4187} & \textbf{0.4512}\\
                               & NMI & 0.5088 & 0.5224 & 0.4700 & 0.5836 & 0.4437 & \underline{0.5852} & \textbf{0.6086}\\
                               & ARI & 0.2273 & 0.2525 & 0.2218 & 0.2687 & 0.2350 & \underline{0.2919} & \textbf{0.3500}\\
    \bottomrule
  \end{tabular}
  \caption{Clustering results of all methods on six datasets. Bold and underline denote the best and second-best results, respectively.}
  \label{tab:clustering}
\end{table*}

\subsection{Specific-to-uniform versus uniform-to-specific}
\label{subsec:3_3}
We discuss the difference between \textit{specific-to-uniform} and \textit{uniform-to-specific} paradigm through the update of fusion parameters $\omega$ with a reconstruction loss $\mathcal{L}_{rec}^v$. Using the same structure described in Sec. \ref{subsec:3_1}, the \textit{specific-to-uniform} generally optimizes $\omega$ by minimizing reconstruction losses over all views, i.e., $\omega^{\ast} = \mathop{\arg\min} \sum_{v=1}^V \mathcal{L}_{rec}^v(r_v(m(z^v, \omega), \phi_{r_v})), x_v)$, and the $\omega$ is updated by (with the SGD)
\begin{equation}
\omega \leftarrow \omega - \alpha \sum_{v=1}^V \nabla_{\omega} \mathcal{L}^v_{rec}(\omega).
\label{eq:S_U_update}
\end{equation}
The optimal $\omega^{\ast}$ observes all views and derives the unified representation $H$, that is, from the particular to the general. While the update of $\omega$ in our \textit{uniform-to-specific} can be written 
\begin{equation}
\omega \leftarrow \omega - \alpha \sum_{v=1}^V \nabla_{\omega} \mathcal{L}^v_{rec} \left(\theta^{\ast}_{b_v}(\omega)\right).
\label{eq:U_S_update}
\end{equation}
Note that $\theta^{\ast}_{b_v}(\omega)$ contain the optimization process of each view in inner-level as Eq. (\ref{eq:inner_update}), which means that the optimal $\omega^{\ast}$ update by observing the reconstruction from unified representation to specific views. Fig. \ref{fig:discuss} intuitively demonstrates the difference between these two manner. 

\subsection{The instances of the objective function}
\label{subsec:3_4}
Our uniform-to-specific framework emphasizes learning from reconstruction in inner-level, thus the $\mathcal{L}_{inner}^v$ is specified as the reconstruction loss \cite{DBLP:conf/cvpr/ZhangLF19, lin2022dual}
\begin{equation}
\mathcal{L}_{inner}^v = \mathcal{L}_{rec}^v(S^v, S_{rec}^v) = \Vert S^v - S_{rec}^v \Vert_F^2.
\label{eq:loss_inner}
\end{equation}
While parameters updated in outer-level can be constrained by richer self-supervised feedbacks as mentained in Sec. \ref{subsec:3_1}. Here we provide two instances of outer-level loss function to demonstrate how MetaViewer can be extended with different learning objectives. 

\textbf{MVer-R} adopts the same reconstruction loss as the inner-level and $\mathcal{L}_{outer} = \sum_{v} \mathcal{L}_{rec}^v(Q^v, Q_{rec}^v)$, which is the purest implementation of MetaViewer.

\textbf{MVer-C} additionally utilizes a contrastive objective, where the similarities of views belonged to same entity (i.e., positive pairs) should be maximized and that of different entities (i.e., negative pairs) should be minimized, i.e., $\mathcal{L}_{outer} = \sum_{v} \left(\mathcal{L}_{rec}^v + \sum_{v^{\prime}, v^{\prime} \neq v}\mathcal{L}_{con}^{v, v^{\prime}}\right)$. Following previous work \cite{DBLP:conf/cvpr/XuT0P0022, DBLP:conf/icml/HassaniA20, DBLP:conf/cvpr/YuLH22}, the contrastive loss $\mathcal{L}_{con}^{v, v^{\prime}}$ is formed as
\begin{equation}
\small
\mathcal{L}_{con}^{v, v^{\prime}} = -\frac{1}{N_{Q}} \sum_{i=1}^{N_{Q}} log\frac{e^{d(q_i^v, q_i^{v^{\prime}})/\tau}}{\sum_{j=1, j\neq i}^{N_{Q}} e^{d(q_i^v, q_j^v)/\tau} + \sum_{j=1}^{N_{Q}} e^{d(q_i^v, q_j^{v^{\prime}})/\tau}},
\label{eq:loss_con}
\end{equation}
where $q_i^v$ is the $v$-$th$ views of the $i$-$th$ query sample, and $d$ is the similarity metric (e.q., cosine similarity \cite{DBLP:conf/icml/ChenK0H20}). The $N_Q$ and $\tau$ denote the number of query set samples and the temperature parameter, respectively. Note that, the derived meta representation can be also used in contrastive learning as a additional noevl view.

\section{Experiments}
\label{subsec:4}
In this section, we present extensive experimental results to validate the quality of the unified representation derived from our MetaViewer. The remainder of the experiments are organized as follows: Subsection \ref{subsec:4_1} lists datasets, compared methods and the implementation details. Subsection \ref{subsec:4_2} compares the performance of our method with classical and state-of-the-art methods on two common downstream scenarios, clustering and classification tasks. Comparison with manually designed fusion and ablation studies are shown in Subsection \ref{subsec:4_3} and \ref{subsec:4_4}, respectively.

\begin{table*}
  \centering
  \begin{tabular}{r|c|cc ccc cc}%{@{}lc@{}}
    \toprule
    Datasets & Metrics  & \makebox[0.08\textwidth][c]{DCCA \cite{DBLP:conf/icml/AndrewABL13}}
                        & \makebox[0.08\textwidth][c]{DCCAE \cite{DBLP:conf/icml/WangALB15}}
                        & \makebox[0.08\textwidth][c]{MIB \cite{DBLP:conf/iclr/Federici0FKA20}}
                        & \makebox[0.08\textwidth][c]{MFLVC \cite{DBLP:conf/cvpr/XuT0P0022}}
                        & \makebox[0.08\textwidth][c]{DCP \cite{lin2022dual}}
                        & \makebox[0.08\textwidth][c]{MVer-R}
                        & \makebox[0.08\textwidth][c]{MVer-C}\\
    \midrule
\multirow{3} * {BDGP}           & ACC       & 0.9840 & \textbf{0.9865} & 0.8900 & 0.9820 & 0.9720 & \underline{0.9860} & 0.9800\\
                                & Precision & 0.9842 & \underline{0.9863} & 0.9005 & 0.9822 & 0.9726 & \textbf{0.9871} & 0.9859\\
                                & F-score   & 0.9840 & \underline{0.9850} & 0.8884 & 0.9820 & 0.9720 & \textbf{0.9859} & 0.9802\\

    \midrule
\multirow{3} * {Handwritten}    & ACC       & 0.8825 & 0.9000 & 0.7900 & 0.9400 & \underline{0.9725} & 0.9700 & \textbf{0.9775} \\
                                & Precision & 0.8920 & 0.9048 & 0.8390 & 0.9420 & \underline{0.9730} & 0.9708 & \textbf{0.9790} \\
                                & F-score   & 0.8805 & 0.8992 & 0.7852 & 0.9401 & \underline{0.9724} & 0.9700 & \textbf{0.9775} \\
    \midrule
\multirow{3} * {RGB-D}           & ACC       & 0.3000 & 0.2400 & 0.3300 & 0.4400 & 0.3700 & \underline{0.5100} & \textbf{0.5600} \\
                                & Precision & 0.2110 & 0.1600 & 0.2850 & 0.4609 & 0.2887 & \textbf{0.5365} & \underline{0.5520} \\
                                & F-score   & 0.2204 & 0.1691 & 0.2737 & 0.4181 & 0.3078 & \underline{0.4873} & \textbf{0.5278} \\

    \midrule
\multirow{3} * {Fashion-MV}      & ACC       & 0.8490 & 0.8535 & 0.8680 & 0.9650 & 0.8925 & \underline{0.9685} & \textbf{0.9770}\\
                                & Precision & 0.8522 & 0.8597 & 0.8680 & \underline{0.9652} & 0.8206 & 0.9637 & \textbf{0.9678}\\
                                & F-score   & 0.8354 & 0.8384 & 0.8655 & \underline{0.9649} & 0.8290 & 0.9648 & \textbf{0.9707}\\
    \midrule
\multirow{3} * {MSRA}           & ACC       & 0.2381 & 0.2429 & 0.3619 & 0.6905 & 0.9048 & \textbf{0.9371} & \underline{0.9270}\\
                                & Precision & 0.2053 & 0.2204 & 0.2498 & 0.7129 & 0.9153 & \textbf{0.9393} & \underline{0.9317}\\
                                & F-score   & 0.2422 & 0.2357 & 0.2773 & 0.6895 & 0.9037 & \textbf{0.9391} & \underline{0.9277}\\
    \midrule
\multirow{3} * {Caltech101-20}  & ACC       & 0.7154 & 0.7154 & 0.7272 & 0.8537 & \textbf{0.9248} & \underline{0.9228} & 0.9216\\
                                & Precision & 0.4527 & 0.6057 & 0.6164 & 0.7183 & 0.8941 & \underline{0.8946} & \textbf{0.9068}\\
                                & F-score   & 0.3981 & 0.4325 & 0.5247 & 0.6907 & \underline{0.8458} & 0.8421 & \textbf{0.8572}\\
    \bottomrule
  \end{tabular}
  \caption{Classification results of all methods on six datasets. Bold and underline denote the best and second-best results, respectively.}
  \label{tab:classification}
\end{table*}

\subsection{Experimental Setup} 
\label{subsec:4_1}

\textbf{Datasets}. To comprehensively evaluate the effectiveness of our MetaViewer, we conduct six multi-view benchmarks in experiments. All datasets are scaled to $[0, 1]$ and split into training, validation, and test sets in the ratio of $6:2:2$, as shown in Tab. \ref{tab:data}. More details of dataset are in Appendix A.

\begin{itemize}
\item \textbf{BDGP} is an image and text dataset, corresponding to $2, 500$ drosophila embryo images in $5$ categories. Each image is described by a $79$-D textual feature vector and a $1, 750$-D visual feature vector \cite{DBLP:journals/bioinformatics/CaiWHD12}.
\item \textbf{Handwritten} contains $2, 000$ handwritten digital images from $0$ to $9$. Two types of descriptors, i.e., $240$-D pixel average in $2\times3$ windows and $216$-D profile correlations, are selected as two views \cite{DBLP:conf/cvpr/ZhangLF19}.
\item \textbf{RGB-D} dataset contains visual and depth images of $300$ distinct objects across $50$ categories \cite{DBLP:journals/corr/abs-1908-01978, DBLP:journals/corr/abs-2204-12496}. Two views are obtained by flattening the $64\times64\times3$ color images and $64\times64$ depth images.
\item \textbf{Fashion-MV} is an image dataset that contains $10$ categories with a total of $30, 000$ fashion products.  It has three views and each of which consists of 10,000 gray images sampled from the same category \cite{DBLP:journals/corr/abs-2103-15069}.
\item \textbf{MSRA} \cite{DBLP:conf/cvpr/XuHN16} consists of $210$ scene recognition images from seven classes with six views, that is, CENTRIST, CMT, GIST, HOG, LBP, and SIFT.

\item \textbf{Caltech101-20} is a subset of the Caltech101 image set \cite{DBLP:journals/cviu/Fei-FeiFP07}, which consists of $2,386$ images of $20$ subjects. Six features are used, including Gabor, Wavelet Moments, CENTRIST, HOG, GIST, and LBP.
\end{itemize}

\textbf{Compared methods.}
We compare the performance of MetaViewer with five representative multi-view learning methods, including two classical methods (DCCA \cite{DBLP:conf/icml/AndrewABL13} and DCCAE \cite{DBLP:conf/icml/WangALB15}) and three state-of-the-art methods (MIB \cite{DBLP:conf/iclr/Federici0FKA20}, MFLVC \cite{DBLP:conf/cvpr/XuT0P0022} and DCP \cite{lin2022dual}). Among them, DCCA and DCCAE are the deep extensions of traditional correlation strategies. MIB is a typical generative method with mutual information constraints. DCP learns unified representation both in complete and incomplete views. In particular, MFLVC also notices the view-private redundant information and designs a multi-level features network for clustering tasks.

\textbf{Implementation details.}
For a fair comparison, all methods are trained from scratch and share the same backbone listed in Appendix B. We concatenate latent features of all views in compared methods to obtain the unified representation $H$ with the same dimension $d_h = {256}$, and verify their performance in clustering and classification tasks using K-means and a linear SVM, respectively. For MetaViewer, we train $2, 000$ epochs for all benchmarks, and set the batch size is $32$ for RGBD and MSRA and $256$ for others. The learning rates in outer- and inner-level are set to $10^{-3}$ and $10^{-2}$, respectively. All experiments have been verified using the PyTorch library on a single RTX3090.

\subsection{Performance on downstream tasks} 
\label{subsec:4_2}
\textbf{Clustering results}. 
Tab. \ref{tab:clustering} lists the results of the clustering task, where the performance is measured by three standard evaluation metrics, i.e., Accuracy (ACC), Normalized Mutual Information (NMI), and Adjusted Rand Index (ARI). A higher value of these metrics indicates a better clustering performance. It can be observed that (1) our MVer-C variant significantly outperforms other compared methods on all benchmarks except MSRA; (2) the second-best results appear between MVer-R and MFLVC, both of which explicitly separates the view-private information; (3) A larger number of categories and views is the main reason for the degradation of clustering performance, and our Metaviewer improves most significantly in such scenario (e.g. Fashion-MV and Caltech101-20). 

\textbf{Classification results}. 
Tab. \ref{tab:classification} lists the results of the classification task, where three common metrics are used including Accuracy, Precision, and F-score. A higher value indicates a better classification performance. Similar to the clustering results, two variants of MetaViewer significantly outperform the comparison methods. It is worth noting that (1) DCP learns the unified generic representation and therefore achieves the second-best result instead of MFLVC. (2) The number of categories is the main factor affecting the classification performance, and our method obtains the most significant improvement in the RGB-D dataset with $50$ classes. More results including incomplete views are deferred to Appendix C.

\subsection{Comparison with manually designed fusion.} 
\label{subsec:4_3}

As mentioned in \ref{subsec:3_3}, MetaViewer is essentially learned to learn an optimal fusion function that filters the view-private information. To verify this, we compare it with commonly used fusion strategies \cite{DBLP:journals/tkde/LiYZ19}, including \textit{sum}, \textit{maxima}, \textit{concatenation}, \textit{linear layer} and \textit{C-Conv}. The former three are the specified fusion rules without trainable parameters and the remaining two are the trainable fusion layer trained via the \textit{specific-to-uniform} manner. Tab. \ref{tab:fusion} lists the clustering results and an additional MSE score on the Handwritten dataset with the same embedding and reconstruction network (see Tab. \ref{tab:data}). We can observe that (1) trainable fusion layers outperform the hand-designed rules, and our MetaViewer yields the best performance; (2) the MSE scores listed in the last column indicate that the quality of the unified representation cannot be measured and guaranteed only with the reconstruction constraint, due to the view-private redundant information mixed in view-specific latent features. 

\begin{figure}[t]
  \centering
  \includegraphics[width=1.\linewidth]{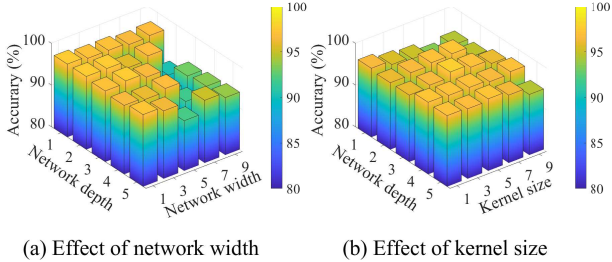}
  \caption{Effect of meta-learner architectures with different depth, width, and kernel size on classification accuracy.}
  \label{fig:kernels}
\end{figure}

\subsection{Ablation Studies}
\label{subsec:4_4}

\textbf{Meta-learner structures}. 
We implement the meta-learner as channel-level convolution layers in this work. Albeit simple, this layer can be considered as a universal approximator for almost any continuous function \cite{DBLP:conf/nips/ShuXY0ZXM19}, and thus can fit a wide range of conventional fusion functions. To investigate the effect of network depth, width, and convolution kernel size on the performance of the representation, we alternate fix the $32$ kernels and $1\times3$ kernel size and show the classification results on Handwritten data in Fig. \ref{fig:kernels}. It is clear that (1) the meta-learner works well with just a shallow structure as shown in Fig. \ref{fig:kernels} (a), instead of gradually overfitting to the training data as the network deepens or widens, (2) our MetaViewer is stable and insensitive to the hyper-parameters within reasonable ranges.

\begin{figure}[t]
  \centering
  \includegraphics[width=1.\linewidth]{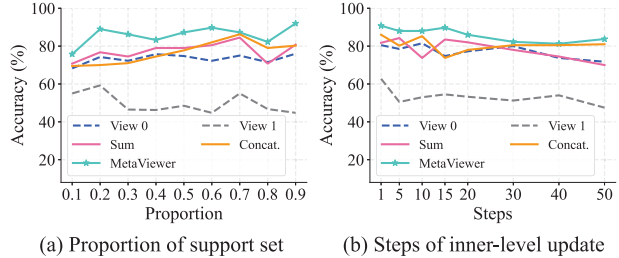}
  \caption{Effect of (a) different meta-division ratios and (b) the number of inner-loop iterations on classification accuracy.}
  \label{fig:inner_ratios}
\end{figure}

\textbf{Meta-split ratios}. 
Fig. \ref{fig:inner_ratios} (a) shows the impact of meta-split mentioned in \ref{subsec:3_2} on the classification performance, where the proportion of support set is set from $0.1$ to $0.9$ in steps of $0.1$, and the rest is query set. In addition to the single view, we also compare the \textit{sum} and \textit{concat.} fusion as baselines. MetaViewer consistently surpasses all baselines over the experimental proportion. In addition, fusion baselines are more dependent on the better-performing view at lower proportions, instead becoming unstable as the available query sample decreases.

\textbf{Inner-level update steps}. 
Another hyper-parameter is the number of iteration steps in inner-level optimization. More iterations mean a larger gap from the learned meta representation to the specific view space, i.e., coarser modeling of view-private information. Fig. \ref{fig:inner_ratios} (b) shows the classification results with various steps, where $n$ step means that the inner-level optimization is updated $n$ times throughout the training. MetaViewer achieves the best results when using $1$ steps, and remains stable within $15$ steps.

\begin{table}
  \centering
  \footnotesize
  \begin{tabular}{c|c|ccc|c}%{@{}lc@{}}
    \toprule
     Strategies & Rules & ACC$\uparrow$ & NMI$\uparrow$ & ARI$\uparrow$ & MSE$\downarrow$\\
    \midrule
    Sum & $z^x + z^y$                   & 69.25 & 71.89 & 59.02 & 1.84\\
    Max & $max(z^x, z^y)$               & 80.75 & 73.93 & 63.75 & -\\
    Concat. & $cat[z^x, z^y]$           & 78.75 & 72.02 & 61.52 & \textbf{1.77}\\
    Linear & $l(z^x, z^y, \theta_l)$    & 85.00 & 77.40 & 69.71 & 4.74\\
    C-Conv & $m(z^x, z^y, \omega)$      & 69.75 & 65.21 & 51.33 & 2.37\\        % 80.5, 75.85, 64.77
    MetaViewer & \textit{meta-learning} & \textbf{86.25} & \textbf{78.96} & \textbf{72.25} & 2.45\\
    \bottomrule
  \end{tabular}
  \caption{Clustering resulting on the Handwritten dataset.}
  \label{tab:fusion}
\end{table}

\section{Conclusion}
\label{subsec:4}
This work introduced a novel meta-learning perspective for multi-view learning, and proposed a meta-learner, namely MetaViewer, to derive a high-quality unified representation for downstream tasks. In contrast to the prevailing \textit{specific-to-uniform} pipeline, MetaViewer observes the reconstruction process from unified representation to specific views and essentially learns an optimal fusion function that separates and filters out meaningless view-private information. Extensive experimental results on clustering and classification tasks demonstrate the performance of the unified representation we meta-learned.

{\small
\bibliographystyle{ieee_fullname}
\bibliography{egbib}
}

\end{document}